\documentclass{article}

\usepackage{microtype}
\usepackage{graphicx}
\usepackage{subfigure}
\usepackage{booktabs} %

\usepackage{hyperref}

\usepackage[accepted]{icml2021}

\usepackage{todonotes}
\usepackage{amsmath}
\usepackage{pseudo}
\usepackage{xspace}
\usepackage{stmaryrd}
\usepackage{amsthm}
\usepackage{multirow}
\usepackage{pifont}%
\usepackage{nicefrac}
\usepackage{wrapfig}
\usepackage{tabularx}
\usepackage{commath}
\usepackage{tikz}
\usetikzlibrary{calc}

\usepackage[compact]{titlesec}  %

\def\equationautorefname~#1\null{Equation~(#1)\null}

\newcommand{\parabf}[1]{\smallskip\noindent\textbf{#1.}}

\usepackage{xspace}
\usepackage{amsfonts}

\newcommand{\ncm}{\ensuremath{A}\xspace}
\newcommand{\score}{\ensuremath{\alpha}\xspace} %
\newcommand{\pval}{\ensuremath{p}\xspace}
\newcommand{\predset}{\ensuremath{\Gamma}\xspace}
\newcommand{\objspace}{\ensuremath{X}\xspace}
\newcommand{\labelspace}{\ensuremath{Y}\xspace}
\newcommand{\nlabels}{\ensuremath{\ell}\xspace}
\newcommand{\prob}[1]{\ensuremath{Pr\left(#1\right)}}

\newcommand{\bigoh}[1]{\ensuremath{\mathcal{O}(#1)\xspace}}
\newcommand{\timetrain}[2][\ncm]{\ensuremath{T_#1(#2)\xspace}}
\newcommand{\timescore}[2][\ncm]{\ensuremath{P_#1(#2)\xspace}}
\newcommand{\timekernel}{\ensuremath{P_K\xspace}}

\newcommand{\cmark}{\ding{51}}%
\newcommand{\xmark}{\ding{55}}%
\newcommand\card{\text{\ttfamily\#}} %
\newcommand{\suchthat}{\ensuremath{\,:\,}} %

\icmltitlerunning{Exact Optimization of Conformal Predictors via Incremental and Decremental Learning}

\begin{document}

\twocolumn[
\icmltitle{Exact Optimization of Conformal Predictors\\
           via Incremental and Decremental Learning}

\author{%
  Giovanni Cherubin\\
  Alan Turing Institute\\
  \texttt{gcherubin@turing.ac.uk}\\
  \And
  Konstantinos Chatzikokolakis\\
  University of Athens\\
  \texttt{kostas@chatzi.org}\\
  \And
  Martin Jaggi\\
  EPFL\\
  \texttt{martin.jaggi@epfl.ch}\\
}

\icmlsetsymbol{equal}{*}

\begin{icmlauthorlist}
\icmlauthor{Giovanni Cherubin}{turing}
\icmlauthor{Konstantinos Chatzikokolakis}{uoa}
\icmlauthor{Martin Jaggi}{epfl}
\end{icmlauthorlist}

\icmlaffiliation{turing}{Alan Turing Institute, London, UK}
\icmlaffiliation{uoa}{University of Athens}
\icmlaffiliation{epfl}{EPFL}

\icmlcorrespondingauthor{Giovanni Cherubin}{gcherubin@turing.ac.uk}

\icmlkeywords{Machine Learning, Conformal Prediction, Optimization}

\vskip 0.3in
]

\printAffiliationsAndNotice{}  %

\begin{abstract}
Conformal Predictors (CP) are wrappers around ML models,
providing error guarantees under weak assumptions on the data distribution.
They are suitable for a wide range of problems, from
classification and regression to anomaly detection.
Unfortunately, their very high computational complexity limits their
applicability to large datasets.
In this work, we show that it is possible to speed up
a CP classifier considerably, by studying it in conjunction with the underlying ML method,
and by exploiting incremental\&decremental learning.
For methods such as k-NN, KDE, and kernel LS-SVM, our
approach reduces the running time by one order of magnitude,
whilst producing exact solutions.
With similar ideas, we also achieve a linear
speed up
for the harder case of
bootstrapping.
Finally, we extend these techniques to improve upon
an optimization of k-NN CP for regression.\\
We evaluate our findings empirically, and discuss
when methods are suitable
for CP optimization.
\end{abstract}

\section{Introduction}
Conformal prediction refers to a set of techniques providing error
guarantees on the predictions of an ML algorithm~\cite{vovk2005algorithmic}.
Its increasing popularity is due to the fact that these
guarantees do not require strict assumptions on the underlying data
distribution; one only needs to assume that the observed examples
are exchangeable (i.e., any permutation of them is equally likely to appear)
-- a weaker requirement than IID.
These guarantees hold for any desired ML algorithm, even if underspecified
or overfitting.

A conformal predictor (CP) can be instantiated for various tasks:
classification and regression~\cite{vovk2005algorithmic},
anomaly detection~\cite{laxhammar2010conformal}, and clustering~\cite{cherubin2015conformal}.
Furthermore, they can be used to test if data is exchangeable (or IID)~\cite{vovk2003testing}.
Our work focuses on classification, and it can be directly applied
to tasks such as anomaly detection, clustering,
and sequence prediction (\autoref{sec:discussion}).
We discuss CP regression separately, in~\autoref{sec:regression}.

\begin{table*}
	\caption{Time complexity of the optimized (our contribution) and
		standard nonconformity measures used for full CP classification. %
		Complexities refer to an $\nlabels$-label classification setting,
		with $n$ training and $m$ test examples.
		Standard full CP requires no training.}
	\label{tab:ncm-complexity-comparison}
	\centering
	\begin{tabular}{lcccc}
		\toprule
		Full CP & & Train & Predict & Exact optimization \\
		\midrule
		(Simplified) k-NN  & Standard & $\bigoh{1}$ & $\bigoh{n^2\nlabels m}$ &  \\
		& Optimized & $\bigoh{n^2}$   & $\bigoh{n\nlabels m}$ & \cmark \\
		\midrule
		KDE   & Standard & $\bigoh{1}$ & $\bigoh{\timekernel n^2\nlabels m}$ \\
		& Optimized & $\bigoh{\timekernel n^2}$ &$\bigoh{\timekernel n\nlabels m}$ & \cmark \\
		&\multicolumn{3}{l}{\scriptsize{$\timekernel$: time complexity of computing kernel $K$ for 1 point}}\\
		\midrule
		LS-SVM   & Standard & $\bigoh{1}$ & $\bigoh{n^{\omega+1}\nlabels m}$ & \\
		& Optimized & $\bigoh{n^\omega}$ & $\bigoh{q^3n\nlabels m}$ & \cmark \\
		&\multicolumn{3}{l}{\scriptsize{$q$: dimensionality of feature vector $\phi(x)$}. For $\omega \in [2, 3]$, $n^\omega$ is the training cost of an LS-SVM model.}\\
		\midrule
		Bootstrap   & Standard & $\bigoh{1}$ & $\bigoh{S_g(n)Bn\nlabels m}$ \\
		& Optimized & $\bigoh{S_g(n)e^{-1}Bn}$ & $\bigoh{S_g(n)(1-e^{-1})Bn\nlabels m}$ & \xmark \\
		&\multicolumn{4}{l}{\scriptsize{
				$B$: n. classifiers}}\\
		&\multicolumn{4}{l}{\scriptsize{
				$S_g(n) = \timetrain[g]{n} + \timescore[g]{1}$: time to train base classifier on $n$ examples
				and make one prediction}}\\
		\bottomrule
	\end{tabular}
\end{table*}

In this paper, we consider the original definition of CP
(also referred to as ``full'' or transductive CP), which is known
to have a good predictive power and to attain the desired
coverage intervals.
Unfortunately, full CP requires running
a leave-one-out (LOO) procedure on the entire training set
for every test point.
This makes
its complexity prohibitive for most real world cases:
if training the ML method on $n$ examples takes $T(n)$ time,
the cost of a CP prediction for $m$ test points is proportional to
$\bigoh{T(n)n\nlabels m}$, for a training set of $n$
examples in an $\nlabels$-label classification setting.
A number of time-efficient modifications of CP exist \cite{vovk2005algorithmic,vovk2015cross,carlsson2014aggregated,barber2019predictive},
which although have a weaker predictive power
and/or coverage guarantees (e.g., \citet{linusson2014efficiency}).

In this work, we focus on exact optimizations of the full CP classification algorithm.
We first observe that,
while a CP can be constructed around virtually any ML method,
most applications of CP classification only use a handful of models.
It therefore makes sense to optimize the CP routine in conjunction
with its underlying model.
In this paper,
we use this idea and exploit
incremental\&decremental learning principles to
produce \textit{exact optimizations} of CP
for: i) k-NN,
ii) Kernel Density Estimation (KDE),
iii) kernel Least-Squares SVM (LS-SVM); all these reduce the complexity by at least one order of magnitude.
Furthermore, iv) we show that bootstrapping methods
can be marginally improved by similar ideas,
and v) we extend our optimizations to CP regression.
Our results demonstrate that full CP is practical for several choices of underlying methods.

\subsection{Related work}

We first review computationally-efficient alternatives to CP,
and then discuss related work on
full CP optimization.

\parabf{Alternatives to full CP}
Despite the desirable properties of full CP, its computational complexity
makes it impractical for most applications.
Researchers have therefore
been investigating modifications of CP, to reduce the computational
complexity.
For example, Inductive CP (ICP), also referred to as ``split CP'', trains the
underlying ML method only on part of the training set, which enables
it to avoid the costly LOO procedure of full CP;
however, this has an impact on its prediction power
(e.g., \autoref{sec:mnist}).
Several methods were proposed after ICP, such as cross-CP~\cite{vovk2015cross},
aggregated CP~\cite{carlsson2014aggregated}, CV+ and the jackknife+~\cite{barber2019predictive}.
These methods mitigated ICP's statistical inefficiency, whilst preserving
a good computational complexity.
However, they have
a weaker prediction power than the
full CP formulation~\cite{linusson2014efficiency, carlsson2017comparing,lei2018distribution,barber2019predictive}.
It is therefore important to have access to efficient optimizations
of full CP, for applications with strict requirements on statistical efficiency
(e.g., \citet{lei2019fast}).

In our experiments, we use ICP as a time complexity baseline for our optimizations,
since it is the most computationally efficient among the above techniques.
We report the time complexity of the other methods in \autoref{appendix:cp-icp-details}.

\parabf{Optimization of full CP classifiers}
A CP is built for an ML method, by converting
the method into a scoring function, the
\textit{nonconformity measure}.
Informally, this function
quantifies the strangeness of an example w.r.t.
training data.

\citet{makili2013incremental} optimized CP
by defining a nonconformity measure based of the
Lagrangian multipliers of a trained SVM.
Thanks to this, they could use
an incremental version of SVM
to avoid the LOO step in CP.
Unfortunately,
this is only a special case of SVM nonconformity measure,
and being incremental
is not sufficient to optimize CP in general:
as we observe in this paper, in order to optimize CP,
an ML method must be both incremental
and decremental.

\citet{vovk2005algorithmic}  optimized CP with the k-NN
nonconformity measure
for online learning settings when parameter $k$ increases
slowly with $n$; they achieved an impressive
$\bigoh{\log(n)}$ time for 1 prediction given $n$ training points.
This method is limited to the Euclidean metric
on $\objspace = [0,1]$,
or
contingent on embedding the object space $\objspace$ in $[0,1]$.
Our k-NN CP optimization works for any
metric space, by exploiting a simple incremental\&decremental
version of k-NN we devise.
Additionally, we show our idea can be used to optimize
KDE CP.

\citet{vovk2005algorithmic} noticed that a linear LS-SVM
nonconformity measure
can be computed efficiently in the LOO step.
In our work, we use the
incremental\&decremental LS-SVM
by \citet{lee2019exact} to generalize this
to multiple kernels.

\parabf{CP regression}
The regression task in CP has been traditionally tackled separately from classification.
In regression, one needs to reformulate CP (and ICP)
to support an infinite label space.
For ICP, this is straightforward and efficient~\cite{papadopoulos2002inductive}.
Other CP modifications for regression exist, e.g.,
conformal predictive distributions \cite{vovk2017nonparametric},
jackknife+, CV+ \cite{barber2019predictive}.
As for full CP, regression is a harder goal, which was achieved only for:
ridge regression~\cite{nouretdinov2001ridge},
k-NN~\cite{papadopoulos2011regression},
and the Lasso~\cite{lei2019fast}.
In \autoref{sec:regression}, by using incremental\&decremental
learning, we produce an exact optimization of the k-NN CP regressor.

\parabf{Contributions}
To summarize our contributions:\\
\vspace{-2.5em}
\begin{itemize}
	\itemsep-0.3em 
	\item We introduce exact optimizations of full CP for the following methods:
			k-NN, ``simplified'' k-NN, KDE, and kernel LS-SVM. Each improves
			at least by one order of magnitude the original complexity (\autoref{tab:ncm-complexity-comparison}).
	\item We further use the incremental\&decremental learning idea to
		optimize bootstrap CP by a linear factor.
	\item We empirically compare our techniques with i) original implementations of full CP, and
		ii) the most computationally efficient CP modification, ICP.
	\item We extend our ideas to CP regression.
		In particular, we improve on an optimization of the k-NN CP regressor
		by \citet{papadopoulos2011regression}, and reduce its
		time complexity
		from $\bigoh{n^2m}$ to $\bigoh{n\log(2n)m}$,
		for predicting $m$ test objects given $n$ training points.
	\item Discuss further optimization avenues for full CP.
\end{itemize}
\vspace{-1em}
\textit{Code to reproduce the experiments:
	\url{https://github.com/gchers/exact-cp-optimization}.}

\section{Preliminaries}
\label{sec:preliminaries}

Consider an $\nlabels$-label classification setting, where
we are given a training set of examples
$Z = \{(x_1, y_1), ..., (x_{n}, y_{n})\} \in (\objspace \times \labelspace)^n$
and we are asked to predict the label for a test object $x$.

We build a \textit{nonconformity measure} on top of an ML method,
as described in \autoref{sec:ncm}.
A nonconformity measure is a real-valued function
$\ncm: (\objspace \times \labelspace) \times (\objspace \times \labelspace)^n \rightarrow \mathbb{R}$,
which quantifies how much an example $(x, y)$ ``conforms to''
(or is similar to) a set
of training examples $\{(x_i, y_i)\}_{i=1}^n$.

For a chosen \textit{significance level} $\varepsilon \in [0,1]$
and nonconformity measure $\ncm$,
a CP classifier
outputs a set
$\predset^\varepsilon \subseteq \labelspace$ as its prediction for test point $x$.
CP guarantees that
this prediction set contains the correct label $y$
with at least $1-\varepsilon$ probability.
Formally, if the set
$\{(x_i, y_i)\}_{i=1}^n \cup \{(x, y)\}$
is exchangeable then
$\prob{y \notin \predset^\varepsilon} \leq \varepsilon$~\cite{vovk2005algorithmic}. %

Because a bound $\varepsilon$ on the probability of error $\prob{y \notin \predset^\varepsilon}$
is chosen in advance,
an analyst only needs to assert that $\predset^\varepsilon$
is statistically \textit{efficient} (i.e., it contains one or very few labels).
The underlying ML method serves this purpose: the better $\ncm$
is, the more efficient %
the prediction set $\predset^\varepsilon$ will be.

In the remainder of this section, we describe how to obtain nonconformity measures from popular ML
methods, %
we outline the CP algorithm and its complexity, and
describe ICP, the computationally-ideal baseline for our optimizations.

\subsection{Nonconformity measures}
\label{sec:ncm}

We call $\ncm$'s output a \textit{nonconformity score};
it takes a smaller value if example $(x, y)$ conforms more
to the training set.
We give two examples of nonconformity
measures.

\parabf{Nearest neighbor}
Let $d$ be a metric on $\objspace$.
The Nearest Neighbor (NN) nonconformity measure is:
\begin{equation}
	\label{eq:nn-ncm}
	\ncm((x, y); \{(x_i, y_i)\}_{i=1}^n) =
	\frac{\min_{i=1, ..., n: y_i=y} d(x, x_i)}
	{\min_{i=1, ..., n: y_i\neq y} d(x, x_i)} \,.
\end{equation}
It is useful to think of a nonconformity measure as a scoring
function determining how suitable label $y$ is for an object $x$;
note that this is equivalent to determining the
conformity of the pair $(x, y)$ to the training data.
The NN nonconformity measure
takes low values if the nearest neighbor to $x$ that has
label $y$ is closer than its nearest neighbor with label
different from $y$; it takes a high value otherwise.
We discuss extensions of this measure in \autoref{sec:knn}.

\parabf{Nonconformity measure from generic ML methods}
Let $f: \objspace \mapsto [0, 1]^{\nlabels}$ be a classifier
returning a confidence score for each of the $\nlabels = |\labelspace|$ labels.
We can construct a nonconformity score from $f$ as follows:
\begin{equation*}
	\label{eq:confidence-classifier-ncm}
	\ncm((x, y); \{(x_i, y_i)\}_{i=1}^n) = -f^y(x) \,,
\end{equation*}
where $f$ is trained on
$\{(x_i, y_i)\}_{i=1}^n$ and
$f^y(x)$ is its score for label $y$.
The negative sign ensures that $\ncm$ takes a lower value if the classifier believes
$y$ is an appropriate label for $x$.

\subsection{Full CP classifier}
\vspace{-2mm}

\begin{algorithm}
\begin{pseudo}
	\pr{compute\_pvalue}(x, \hat{y}, \ncm, Z=\{(x_i, y_i)\}_{i=1}^n)\\+
		$\score = \ncm((x, \hat{y}); Z)$\\
		for $i$ in 1, ..., n\\+
			$\score_i = \ncm((x_i, y_i); \{(x, \hat{y})\} \cup Z \setminus \{(x_i, y_i)\})$\\-
		$\pval_{(x, \hat{y})} = \frac{\card\{ i=1, ..., n \suchthat \score_i \geq \score\} + 1}{n+1}$\\
		
		return $\pval_{(x, \hat{y})}$
\end{pseudo}
\caption{CP: computing a p-value for $(x, \hat{y})$}
\label{algo:cp}
\end{algorithm}

Let $Z=\{(x_i, y_i)\}_{i=1}^n$ be a training set, and $x$ a
test object.
For each possible label $\hat{y} \in \labelspace$,
CP computes a p-value $\pval_{(x, \hat{y})}$ (\autoref{algo:cp})
based on the hypothesis
that $(x, \hat{y})$ comes from the same distribution as $Z$;
intuitively, $p_{(x, \hat{y})}$ attests on whether
$\hat{y}$ is a good label for $x$.
CP outputs the following set as its prediction:
$\predset^\varepsilon = \{\hat{y}\in\labelspace \suchthat p_{(x, \hat{y})} > \varepsilon\}$,
for a desired value $\varepsilon \in [0, 1]$.

\parabf{Time complexity of CP}
Let $\timetrain{n}$ be the time to train $\ncm$
on a dataset $Z$ of $n$ examples,
and $\timescore{m}$ that of
using the trained $\ncm(\cdot; Z)$ to predict $m$ examples.
\autoref{algo:cp} has complexity
$\bigoh{(\timetrain{n}+\timescore{1})n}$.
If we assume the nonconformity measure should at least inspect every
training point (i.e., $\timetrain{n}=n$),
a lower bound on the complexity
to compute the p-value \textit{for one test point}
is $\bigoh{n^2}$.

When used for classifying a test object $x$ in a set of labels $\labelspace$,
CP needs to run \autoref{algo:cp}
for every possible pairing $(x, \hat{y})$, $\hat{y} \in \labelspace$.
Therefore, the complexity becomes $\bigoh{(\timetrain{n}+\timescore{1})n\nlabels}$,
where $\nlabels = |\labelspace|$.
The lower bound is $\bigoh{n^2\ell}$ for classifying one test point.

\subsection{Inductive CP classifier}

The most computationally-efficient -- alas statistically inefficient,
alternative to CP is inductive CP (ICP)~\cite{vovk2005algorithmic}.
For a parameter $t \in \{1, ..., n\}$,
ICP splits the training set $Z$ into:
\textit{proper} training set $Z_{train}$ and
\textit{calibration} set $Z_{calib}$, where $Z_{train} \cup Z_{calib} = Z$,
and $|Z_{train}| = t$.
Then it trains the nonconformity measure $\ncm$ on $Z_{train}$,
and it computes the scores $\alpha_i = \ncm((x_i, y_i); Z_{train})$
only for the calibration examples $(x_i, y_i) \in Z_{calib}$,
instead of the entire training set; this
avoids the LOO step (Lines 3-4, \autoref{algo:cp}).
ICP is outlined in \autoref{appendix:cp-icp-details}.

\parabf{Time complexity of ICP}
Consider an ICP trained on $n$ examples, $t$ of which are
used for the proper training set.
The running time for training and calibration is
$\bigoh{\timetrain{t}+\timescore{n-t}}$.
The time for computing the p-value for one example is $\bigoh{\timescore{1}+n-t}$.
This becomes $\bigoh{(\timescore{1}+n-t)\nlabels}$ when
classifying one test object into $\nlabels$ labels.

\section{Nearest neighbor nonconformity measures}
\label{sec:knn}
We describe nonconformity measures based on the nearest neighbor principle,
and introduce an optimization for their use in CP.
Let $d$ be a distance metric in the object
space $\objspace$.

\parabf{k-NN}
\autoref{eq:nn-ncm} is the NN nonconformity measure,
measuring the ratio of the smallest distance from examples
with the same label and
examples with a different label.
We study a generalization of this according to the k-NN principle.

Let $\delta^j(x, S)$ be the $j$-th smallest distance of object
$x$ from the points in set $S$. The k-NN measure is~\cite{vovk2005algorithmic}:
\begin{equation}
\begin{split}
\label{eq:knn-ncm}
A((x, y); &\{(x_i, y_i)\}_{i=1}^n) = \\
&\frac{\sum_{j=1}^k \delta^j(x, \{x_i \suchthat i=1...n, y_i=y\})}
{\sum_{j=1}^k \delta^j(x, \{x_i \suchthat i=1...n, y_i\neq y\})} \,.
\end{split}
\end{equation}

\parabf{Simplified k-NN}
Another version of the k-NN nonconformity measure, useful
for anomaly detection \cite{laxhammar2010conformal},
is defined as the nominator of \autoref{eq:knn-ncm}:
$A((x, y); \{(x_i, y_i)\}_{i=1}^n) =
\sum_{j=1}^k \delta^j(x, \{x_i \suchthat i=1...n, y_i=y\})$.
Because it only contains information for one label, we refer
to it as the simplified k-NN measure.

\parabf{Complexity}
CP classification of $m$ test points takes $\bigoh{n^2\nlabels m}$ for
both Simplified k-NN and k-NN.
We report the derivation for all the complexities in
\autoref{appendix:complexity-derivations} and \autoref{appendix:memory-costs}.
They are summarized in \autoref{tab:ncm-complexity-comparison}.

\subsection{Optimizing nearest neighbor CP}

\begin{figure}
\centering
\subfigure[$\alpha_i'$ is updated]{
	\tikzset{every picture/.style={line width=0.75pt}} %
	
	\tikzset{every picture/.style={line width=0.75pt}} %
	
	\begin{tikzpicture}[x=0.75pt,y=0.75pt,yscale=-1,xscale=1]
	
	\draw  [dash pattern={on 0.84pt off 2.51pt}]  (331.5,375.83) -- (310.7,359.3) ;
	\draw [color={rgb, 255:red, 208; green, 2; blue, 27 }  ,draw opacity=1 ] [dash pattern={on 0.84pt off 2.51pt}]  (331.5,296) -- (308.5,342.5) ;
	\draw  [dash pattern={on 0.84pt off 2.51pt}]  (269.5,401.83) -- (296.17,361.83) ;
	
	\draw (284.2,301.5) node [anchor=north west][inner sep=0.75pt]  [font=\footnotesize,color={rgb, 255:red, 208; green, 2; blue, 27 }  ,opacity=1 ]  {$d( x_{i} ,x)$};
	\draw  [fill={rgb, 255:red, 245; green, 166; blue, 35 }  ,fill opacity=1 ]  (301.93, 351.53) circle [x radius= 12.04, y radius= 12.04]   ;
	\draw (301.93,351.53) node  [font=\footnotesize]  {$x_{i}$};
	\draw  [fill={rgb, 255:red, 245; green, 166; blue, 35 }  ,fill opacity=1 ]  (340.33, 382.44) circle [x radius= 11.31, y radius= 11.31]   ;
	\draw (340.33,382.44) node  [font=\footnotesize]  {$$};
	\draw  [fill={rgb, 255:red, 245; green, 166; blue, 35 }  ,fill opacity=1 ]  (261.13, 410.04) circle [x radius= 11.31, y radius= 11.31]   ;
	\draw (261.13,410.04) node  [font=\footnotesize]  {$$};
	\draw  [fill={rgb, 255:red, 245; green, 166; blue, 35 }  ,fill opacity=1 ]  (279.52, 258.84) circle [x radius= 11.31, y radius= 11.31]   ;
	\draw (279.52,258.84) node  [font=\footnotesize]  {$$};
	\draw (277.15,373.6) node  [font=\footnotesize]  {$\Delta _{i}^{2}$};
	\draw (327.15,359.1) node  [font=\footnotesize]  {$\Delta _{i}^{1}$};
	\draw  [fill={rgb, 255:red, 255; green, 255; blue, 255 }  ,fill opacity=1 ]  (336.53, 286.34) circle [x radius= 11.31, y radius= 11.31]   ;
	\draw (336.53,286.34) node  [font=\footnotesize]  {$x$};

	\end{tikzpicture}
	
}
\hfill
\subfigure[No update]{
	\tikzset{every picture/.style={line width=0.75pt}} %
	
	\begin{tikzpicture}[x=0.75pt,y=0.75pt,yscale=-1,xscale=1]
	
	\draw  [dash pattern={on 0.84pt off 2.51pt}]  (173,376.33) -- (152.2,359.8) ;
	\draw  [dash pattern={on 0.84pt off 2.51pt}]  (123.5,270.5) -- (139.5,341.5) ;
	\draw  [dash pattern={on 0.84pt off 2.51pt}]  (111,402.33) -- (137.67,362.33) ;
	
	\draw  [fill={rgb, 255:red, 245; green, 166; blue, 35 }  ,fill opacity=1 ]  (143.43, 352.03) circle [x radius= 12.04, y radius= 12.04]   ;
	\draw (143.43,352.03) node  [font=\footnotesize]  {$x_{i}$};
	\draw  [fill={rgb, 255:red, 245; green, 166; blue, 35 }  ,fill opacity=1 ]  (181.83, 382.94) circle [x radius= 11.31, y radius= 11.31]   ;
	\draw (181.83,382.94) node  [font=\footnotesize]  {$$};
	\draw  [fill={rgb, 255:red, 245; green, 166; blue, 35 }  ,fill opacity=1 ]  (102.62, 410.54) circle [x radius= 11.31, y radius= 11.31]   ;
	\draw (102.62,410.54) node  [font=\footnotesize]  {$$};
	\draw  [fill={rgb, 255:red, 245; green, 166; blue, 35 }  ,fill opacity=1 ]  (121.03, 259.34) circle [x radius= 11.31, y radius= 11.31]   ;
	\draw (121.03,259.34) node  [font=\footnotesize]  {$$};
	\draw (144.15,303.1) node  [font=\footnotesize]  {$\Delta _{i}^{3}$};
	\draw (118.65,374.1) node  [font=\footnotesize]  {$\Delta _{i}^{2}$};
	\draw (168.65,359.6) node  [font=\footnotesize]  {$\Delta _{i}^{1}$};
	\draw  [fill={rgb, 255:red, 255; green, 255; blue, 255 }  ,fill opacity=1 ]  (188.03, 251.34) circle [x radius= 11.31, y radius= 11.31]   ;
	\draw (188.03,251.34) node  [font=\footnotesize]  {$x$};

	\end{tikzpicture}
}

\caption{Intuition behind the Simplified k-NN optimization.
	Training points: \tikz[baseline=(char.base)]{
		\node[shape=circle, draw, inner sep=1pt, 
		minimum height=12pt,fill={rgb, 255:red, 245; green, 166; blue, 35}] (char) {\vphantom{1g}};},
	test point: 	\tikz[baseline=(char.base)]{
		\node[shape=circle, draw, inner sep=1pt, 
		minimum height=12pt] (char) {\vphantom{1g}$x$};};
	$k=3$.
	The nonconformity score $\alpha_i$ for training point $x_i$
	only depends on its $k$ closest points.
	The provisional nonconformity score $\score_i'$ is
	updated if test point $x$ is a $k$-NN of $x_i$ (a);
	otherwise, no update occurs: $\alpha_i = \alpha_i'$ (b).
}
\label{fig:nn-update}
\end{figure}

The bottleneck of \autoref{algo:cp} is computing the nonconformity score for
each training example,
$\score_i = \ncm((x_i, y_i); \{(x, y)\} \cup Z \setminus \{(x_i, y_i)\}) \,,$
where $Z$ is the training set.
We observe that, in order to speed this up, the nonconformity measure should be able to
efficiently both learn a new
example (the test example), and unlearn an example (the $i$-th example in
the LOO step).
That is, we need to devise an incremental\&decremental version of k-NN.

To this end, we get inspiration from classical techniques for LOO k-NN cross validation
(e.g. \citet{fukunaga1989leave,hamerly2010efficient}), although
these are not directly applicable to our setting.
The main difference is that in CP we can
precompute the distances that are subsequently used to predict
a test point; this enables improving the performance further.

We focus on optimizing Simplified k-NN, although the same arguments
apply to k-NN.
Our proposal is based on the observation that
nearest neighbor measures only depend on a subset of
($k$) examples.
We exploit this as follows.
In the training phase, we precompute provisional scores:
$$\score_i' = \ncm((x_i, y_i); Z \setminus \{(x_i, y_i)\}) = \sum_{j=1}^k \Delta^j_i \,,$$
where, for $j=1, ..., k$:
$$ \Delta^j_i = \delta^j(x_i, \{x_a \suchthat (x_a, y_a) \in Z\setminus \{(x_i, y_i)\}, y_a = y_i\}) \,.$$
Scores $\score_i'$ are provisional, because they do not account for
the test example  $(x, y)$.
In the prediction phase,
to compute the p-value for $(x, y)$, we update the
scores as follows:
\begin{equation*}
	\score_i = 
	\begin{cases}
		\score_i' -  \Delta^k_i + d(x_i, x) \quad & \text{if } \Delta^k_i > d(x_i, x) \text{ and } y_i = y\\
		\score_i' \quad &\text{otherwise} \,,
	\end{cases}
\end{equation*}
where $\Delta^k_i$
is the $k$-th smallest distance from $x_i$ to the training examples (excluding $(x_i, y_i)$) with the same label as $x_i$.
That is, we only update score $\score_i$, associated with
$(x_i, y_i)$, if $(x, y)$ is among its $k$ nearest neighbors.
This is illustrated in \autoref{fig:nn-update}.
The cost is $\bigoh{1}$.

The k-NN measure is optimized similarly,
by keeping for each training example its $k$ best distances from
both objects with the same label and from those with a different label.

\parabf{Complexity}
For both measures, the training cost is $\bigoh{n^2}$.
Classifying $m$ test examples is $\bigoh{n\nlabels m}$.

\section{Kernel Density Estimation}
For a kernel function $K$,
the Kernel Density Estimation (KDE) nonconformity measure is:
\begin{equation*}
\ncm((x, y); \{(x_i, y_i)\}_{i=1}^n) =
- \frac{1}{n_yh^p} \sum_{x_i: y_i = y} K \left(\frac{x - x_i}{h}\right) \,,
\end{equation*}

\vspace{-0.2cm}
where $n_y = \card\{ i=1, ..., n \suchthat y_i = y \}$,
$h$ is the bandwidth, and $p$ is the objects' dimensionality.

\parabf{Complexity}
If computing
the kernel for one object is $\timekernel$,
CP classification takes
$\bigoh{\timekernel n^2\nlabels m}$.

\subsection{Optimizing KDE CP}
We use a similar idea to that of our k-NN optimization;
however, in this case $\ncm$ depends on all the training points, not just a subset.
To the best of our knowledge, this incremental\&decremental adaptation
of KDE is also novel.
For training, we compute preliminary scores:
$$\score_i' = \sum_{x_j: y_j = y_i} K \left(\frac{x_i - x_j}{h}\right) \quad i=1, ..., n \,.$$
To calculate the p-value for an example $(x, y)$ in the test phase,
we update the scores as follows:
\begin{equation*}
\score_i = \begin{cases}
-\frac{1}{n_yh^d}\left(\score_i' + K \left(\frac{x - x_i}{h}\right)\right) \quad &\text{if } y_i=y\\
-\frac{1}{n_yh^d}\score_i' \quad &\text{otherwise} \,.
\end{cases}
\end{equation*}

\vspace{-0.25cm}
\parabf{Complexity} Training takes $\bigoh{\timekernel n^2}$.
CP classification runs in $\bigoh{\timekernel n\nlabels m}$.

\section{Least Squares Support Vector Machine}
\label{sec:svm}

Assume $\labelspace = \{-1, 1\}$.
Consider a feature map $\phi: \objspace \rightarrow F$.
The least-squares SVM (LS-SVM) regressor, defined by $\phi$ and a vector $w$, returns a prediction
for an object $x$ as: $w^\top \phi(x)$.
The model $w$ is trained with Tikhonov regularization (ridge regression); details
in \autoref{appendix:lssvm-bootstrap-details}.
We define the nonconformity measure for the LS-SVM regressor as:
\begin{equation*}
\ncm((x, y); \{(x_1, y_1), ..., (x_{n}, y_{n})\}) = -y f(x) \,;
\end{equation*}
it takes high values if the prediction $f(x)$ is different
(in sign) from $y$.
Extension of this
to $\nlabels = |\labelspace| >2$ %
can be done via one-vs-rest approaches (e.g., \citet{vovk2005algorithmic}).

\vspace{-0.1cm}
\parabf{Complexity} Depending on algorithm choices, training
LS-SVM takes $n^\omega$, $\omega \in [2, 3]$. CP LS-SVM
takes
$\bigoh{n^{\omega+1}\nlabels m}$.

\subsection{Optimizing LS-SVM CP}

We exploit recent work by \citet{lee2019exact}, which enables exact
incremental and decremental learning of LS-SVM.
Given a trained model $w$, their proposal enables
updating $w$ by adding/removing an example in
time $\bigoh{q^3}$, where $q$ is the dimensionality of the
feature space $F$ (\autoref{appendix:lssvm-bootstrap-details}).

We apply this for optimizing LS-SVM CP.
In the training phase, we learn
the model $w$ on the training data.
Then, to compute the nonconformity score for an example $(x_i, y_i)$,
we: i) update the model with the test example by using the approach
by Lee et al.,
ii) make a prediction for $(x_i, y_i)$.

\parabf{Complexity} Training LS-SVM takes $\bigoh{n^\omega}$, for $\omega \in [2, 3]$ (one-off cost).
CP classification is $\bigoh{q^3n\nlabels m}$.

\parabf{Discussion}
Other options are possible for optimizing SVM nonconformity
measures.
\citet{cauwenberghs2001incremental} proposed an incremental\&decremental
version of SVM, which differently from the one we used has a larger memory footprint.
Another promising avenue for optimization is the classical linear SVM formulation using coordinate-descent, in combination with incremental updates~\cite{tsai2014incremental}.

\begin{figure*}[ht!]
	\centering
	\includegraphics[width=0.9\textwidth]{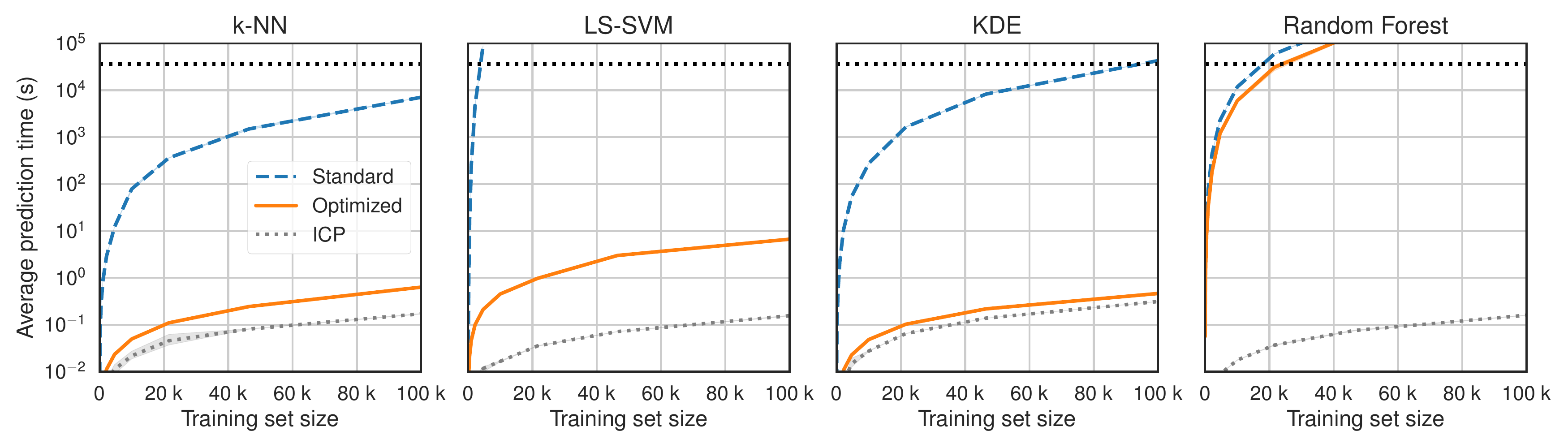}
	\caption{Comparison between the standard and optimized full CP.
		ICP serves as a baseline for these measurements.
		Prediction time for one test point w.r.t. the size of training data.
		Black dashed line is the  experiment timeout (10 hours).
	}
	\label{fig:prediction-time-all}
\end{figure*}

\section{Bootstrapping methods}
\label{sec:bootstrap}

Let integer $B > 1$ be a hyperparameter, and select a \textit{base classifier}
(e.g., decision tree).
In bootstrapping, the training data $Z = \{(x_1, y_1), ..., (x_{n}, y_{n})\}$
is sampled $B$ times with replacement to produce
$B$ bootstrap samples, $Z_1, ..., Z_B$.
On each sample we fit the base classifier,
obtaining an ensemble of $B$ classifiers $(g_1, ..., g_B)$, which we
jointly denote with $f: \objspace \rightarrow [0,1]^{\nlabels}$,
$\nlabels = |\labelspace|$.

Classifier $f$ outputs a confidence vector, $f(x) \in [0,1]^{\nlabels}$,
over the labels.
The $y$-th element of this vector, denoted by $f^y(x)$,
is computed as the normalized count of classifiers $g_i$
that predict $y$. That is:\vspace{-1mm}
$$f^y(x) = \frac{1}{B}\card\{i=1, ..., B \suchthat g_i(x) =  y\} \quad y\in \labelspace \,.$$

We define the  bootstrapping nonconformity measure as:
\begin{equation*}
\ncm((x, y); \{(x_1, y_1), ..., (x_{n}, y_{n})\}) = -f^y(x) \,.
\end{equation*}

\parabf{Complexity}
Let $\timetrain[g]{n}$ be the time needed to train the base
classifier on $n$ training points, and $\timescore[g]{m}$
its cost to predict $m$ points.
Bootstrap CP runs in $\bigoh{(\timetrain[g]{n} + \timescore[g]{1})Bn\nlabels m}$.

\subsection{Optimizing bootstrap CP}

Standard bootstrap CP requires training a bootstrap ensemble for each
training example $(x_i, y_i)$ and one for the test example $(x, y)$;
this entails creating, for each example, $B$ bootstrap samples
that do not contain that example.
The optimization we propose maintains the spirit of bootstrap,
although it may lead to different results from the standard
version because of changes in the sampling strategy.

We first explain the basic idea for training and prediction,
and then improve it with two remarks.
Let ``$\ast$'' be a placeholder for the test point $(x, y)$,
which is unavailable during training, and let
$Z^\ast = Z \cup \{\ast\}$ be the augmented training set.
For a number $B' > B$ to be later specified, we create
$B'$ bootstrap samples of $Z^\ast$, denoted $\{Z^\ast_1, ..., Z^\ast_{B'}\}$.
We continue creating samples until, for every point
$(x_i, y_i) \in Z^\ast$, there are at least $B$ bootstrap samples that
\textit{do not} contain $(x_i, y_i)$;
that is, we increase the number of samples $B'$ until
$\card\{ b=1, ..., B' \suchthat (x_i, y_i) \notin Z^\ast_b\} \geq B$
for all $(x_i, y_i) \in Z^\ast$.
This ensures that each training point (and the placeholder
test point) have at least $B$ bootstrap samples.\footnote{If at the end of the
	procedure an example has more than $B$, we can truncate them
	to $B$ to save up on computational resources.}

Let $E_i = \{ b=1, ..., B' \suchthat (x_i, y_i) \notin Z^\ast_b\}$ be the
samples associated with $(x_i, y_i)$, and
$E = \{ b=1, ..., B' \suchthat \ast \notin Z^\ast_b\}$ the ones associated
with the (placeholder) test example.
In the prediction phase, we compute a prediction for test point
$(x, y)$ by using the base classifiers trained on the
bootstrap samples in $E$.
We make the prediction for a training point $(x_i, y_i)$
in the LOO procedure of CP as follows:
i) in $E_i$'s bootstrap samples, replace the placeholder $\ast$ with the test point $(x, y)$,
ii) train the base classifiers on the samples from $E_i$ and compute
a prediction for $(x_i, y_i)$. %

\parabf{Remarks}
The procedure explained so far preemptively samples $B$ bootstraps
for each point.
We apply the following improvements.
Because some bootstrap samples $Z^\ast_b \in E_i$ associated with
$(x_i, y_i)$ do not contain the placeholder $\ast$, in the
training phase we: i) pretrain the base classifiers $g_b(x)$
on them, and ii) compute their predictions for $(x_i, y_i)$.
This saves up considerable time in the prediction phase.
The optimized bootstrap algorithm is listed in \autoref{appendix:lssvm-bootstrap-details}.

\parabf{Complexity}
Optimized CP classification for $m$ test points is
$\bigoh{(\timetrain[g]{n}+\timescore[g]{1})(1-e^{-1})B\nlabels m}$,
a factor $(1-e^{-1}) \approx 0.632$
speed up on the standard one.
The speed up of this optimization is not as prominent as our
other proposals.
However, we suspect one can further improve bootstrap CP
for base classifiers that support incremental\&decremental learning (\autoref{sec:discussion}).
We leave this to future work.

\section{Empirical evaluation}
\label{sec:experiments}

We compare the running time of the original and optimized CP,
using ICP as a baseline.
We detail hardware, precautions taken to ensure the
fidelity of the measurements, and hyperparameters
in \autoref{appendix:experiment-parameters}.
We instantiate bootstrap CP to Random Forest.

\subsection{Comparison between standard and optimized CP}
\label{sec:experiments-cp}

\parabf{Setup}
In our experiments, the data distribution is irrelevant.
We generate data for a binary classification problem with $30$ features,
by using the \texttt{make\_classification()} routine of the \texttt{scikit-learn} library.
(In \autoref{sec:mnist}, we further compare CP and
ICP on the \texttt{MNIST} dataset.)

For every training size $n$, chosen in the space $[10, 10^5]$,
we train the CP with a nonconformity measure, and use it to predict
$100$ test points.
We set a timeout of 10 hours, which is verified after the prediction of every test
point; therefore, the timeout may be exceeded if the prediction for a point has
already started.
We measure both the training time and the average prediction time for a test point.
Each experiment is repeated for 5 different initialization seeds.

\parabf{Prediction time}
\autoref{fig:prediction-time-all} shows the comparison between standard and optimized CP.
Results confirm the complexity we derived analytically.
For 100k training points,
the optimized k-NN CP  ensures a prediction in 0.63 seconds,
whilst the respective unoptimized version takes roughly 2 hours for the same prediction.
Since k-NN and Simplified k-NN behave very similarly,
results for the latter are in \autoref{appendix:full-results}.
The largest speed up is with LS-SVM:
the optimized version has a running time of 0.21 seconds; the standard implementation
takes on average more than 24.5 hours for 1 prediction.
Our bootstrap CP optimization only gives
a marginal improvement over the original implementation.
For
$n=46415$, optimized Random Forest takes 43 hours for one prediction,
the standard one 82 hours.

\parabf{Comparison with ICP}
We use ICP as a baseline.
For a parameter $t \in \{1, ..., n\}$,
ICP trains the nonconformity measure on a subset of $t$ examples,
and computes the scores for the remaining $n-t$.
We fix $t/n = 0.5$.

As expected, results (\autoref{fig:prediction-time-all}) show that
ICP is strictly faster than the optimized CP methods: e.g., when trained on 100k examples,
LS-SVM takes 6.68 seconds per prediction, while LS-SVM ICP takes 0.16 seconds;
the worst performing is Random Forest, which as seen above improves CP only
by a linear factor.
Nevertheless, in some cases ICP and optimized CP have the same magnitude:
for KDE, ICP takes 0.31 seconds, optimized CP 0.46 seconds.
In other words, our CP optimization seems to perform comparably well to
ICP on reasonably large datasets.

This reveals a better trade-off between computational-statistical efficiency in
conformal inference: if one's priority is speed, they can use ICP, or other CP alternatives; however,
if they can sacrifice computational time, they can get full CP
predictions and yet scale to real-world data.

\subsection{Training time}
\begin{figure}
	\centering
	\includegraphics[width=0.55\linewidth]{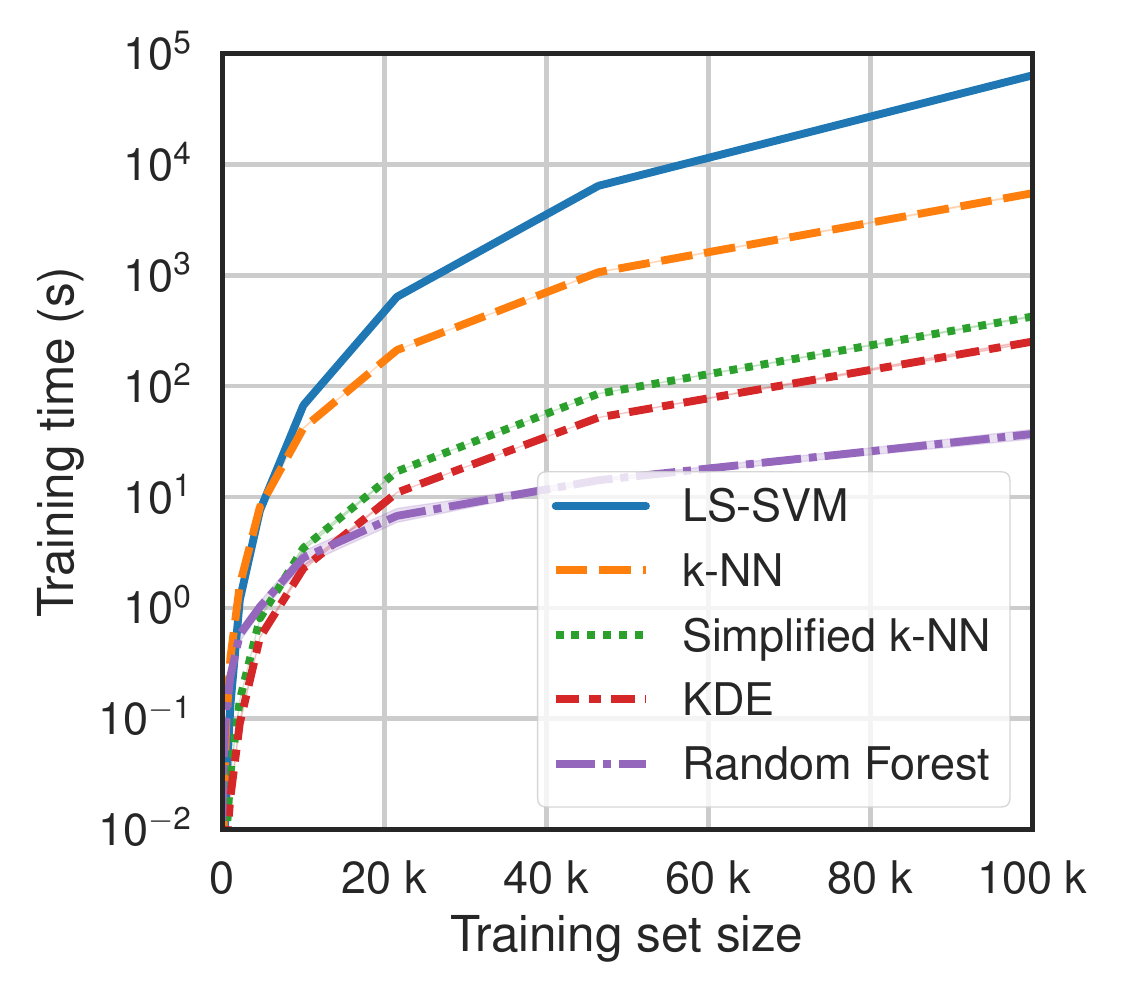}
	\caption{Training time of optimized CP.}
	\label{fig:training-time}
\end{figure}

CP with the optimized nonconformity measures incurs into a
training time, while standard CP does not.
We compare the training time of the optimized measures
in \autoref{fig:training-time}.

We observe that LS-SVM has the highest training time, whilst Random
Forest the lowest.
We also notice that the training time is a reasonable price
to pay in practice.
In a batch classification setting with 100k training
and 20k test examples, optimized k-NN CP would take 2.2 hours for training
and 3.3 hours for prediction.
Standard k-NN CP would have no training time, but its prediction
routine would run for 9.3 years to obtain the same solution.

It may be possible to speed up our techniques even further via
approximate incremental\&decremental learning techniques.
We leave this to future work (\autoref{sec:discussion}).

\section{Large $\labelspace$ and extension to regression}
\label{sec:regression}

The classification algorithms for CP (\autoref{algo:cp}) and ICP (\autoref{algo:icp})
are clearly unfeasible for a very large $\labelspace$:
they both require repeating the calculations for each $y \in \labelspace$.

Things are different for regression, where we assume a total order on $\labelspace$.
In this case, one can avoid the $\nlabels = |\labelspace|$ term in the
cost of both CP and ICP \cite{vovk2005algorithmic}.
Indeed, it is possible to find the intervals of $\labelspace$ where
the p-value $\pval_{(x, \hat{y})}$ exceeds $\varepsilon$, without having
to try all values $\hat{y} \in \labelspace$.
In ICP, this can be done efficiently for general regressors.

As for full CP, this optimization is harder, as one needs to
update the intervals of $\labelspace$ for each training point when
a new point arrives.
Full CP regression was optimized in this
sense for k-NN \cite{papadopoulos2011regression}, ridge regression \cite{nouretdinov2001ridge}, and Lasso \cite{lei2019fast}.
\citet{ndiaye2019computing} recently proposed a general method
leading to approximate but statistically valid CP regressors.

Since the above full CP regression methods do not exploit
incremental\&decremental ideas, we suspect they can be optimized further.
We show this is possible for k-NN.

\subsection{Improving the k-NN CP regressor}

The full k-NN CP regressor
works as follows.
Fix an hyperparameter $k > 0$.
Let $\tilde{y} \in \labelspace$ be a candidate label (not to be defined explicitly)
for test object $x$.
Define the nonconformity score for the $i$-th training example $(x_i, y_i)$ as:
\begin{equation*}
	\alpha_i = \alpha_i(\tilde{y}) = \abs{a_i + b_i\tilde{y}} \,,
\end{equation*}
where, for $i=1, ..., n$:
\begin{align*}
	a_i &= \begin{cases}
		y_i - \frac{1}{k}\sum_{j=1}^{k-1}y_{(j)}(x_i) \quad &\mbox{if $x$ is one of $x_i$'s $k$ NNs}\\
		y_i - \frac{1}{k}\sum_{j=1}^k y_{(j)}(x_i) \quad &\mbox{otherwise} \,,
	\end{cases}\\
	b_i &= \begin{cases}
		-\frac{1}{k} \quad &\mbox{if $x$ is one of $x_i$'s $k$ NNs}\\
		0 \quad &\mbox{otherwise} \,;
\end{cases}
\end{align*}
here $y_{(j)}(x_i)$ is the label of the $j$-th nearest neighbor of $x_i$ in the training set $Z \setminus (x_i, y_i)$.
For the test example $x$, we set $a = -\nicefrac{1}{k} \sum_{j=1}^k y_{(j)}(x)$, $b = 1$.
The p-value is:
\begin{equation*}
	\pval_{(x, \tilde{y})} = \frac{\card\{i=1, ..., n \suchthat \abs{a_i + b_i\tilde{y}} \geq \abs{a + b\tilde{y}}\}}{n+1} \,.
\end{equation*}
The optimization idea by  \citet{papadopoulos2011regression} is based on the fact that,
in order to find an interval of $\labelspace$ for which $\pval_{(x, \tilde{y})} > \varepsilon$,
it suffices to find the points $\tilde{y} \in \labelspace$ for which $c(\tilde{y}) = \alpha_i(\tilde{y}) - \alpha(\tilde{y})$
changes. This can be done efficiently, by only looking at most at $2n$ points.
The time complexity of one prediction is $\bigoh{n^2 + 2n\log(2n)}$, where the
term $\bigoh{n^2}$ comes from computing the $k$ nearest neighbors of each training
point, and $\bigoh{2n\log(2n)}$ comes from sorting the critical points of $c(\tilde{y})$
(required by the above algorithm).

\parabf{Optimization via incremental\&decremental learning}
The method by \citet{papadopoulos2011regression} can be further improved via the incremental\&decremental
k-NN algorithm we proposed in this paper.
We reduce the $\bigoh{n^2}$ term as follows.
In the training phase, we: i) precompute the pairwise distances of
the training points in $Z$, ii) and precompute temporary values
$a_i'$ and $b_i'$, for $i=1, ..., n$. Specifically, we let
$a_i = y_i - \frac{1}{k}\sum_{j=1}^k y_{(j)}(x_i)$ and $b_i = 0$,
as if the (yet unknown) test example $x$ did not contribute to their values.
When making a prediction for $x$, we:
iii) compute its distance from the elements of $Z$ (takes $\bigoh{n}$),
and iv) update those $a_i'$ and $b_i'$ such that $x$ is one of
the $k$- nearest neighbors of $(x_i, y_i)$.
Then we proceed as before.

Even in this setting, using an incremental\&decremental version of
the nonconformity measure enables us to reduce the prediction complexity
by almost one order of magnitude.
Predicting $m$ test examples reduces from $\bigoh{(n^2 + 2n\log(2n))m}$
to $\bigoh{2n\log(2n)m}$.

\begin{figure}
	\centering
	\includegraphics[width=0.55\linewidth]{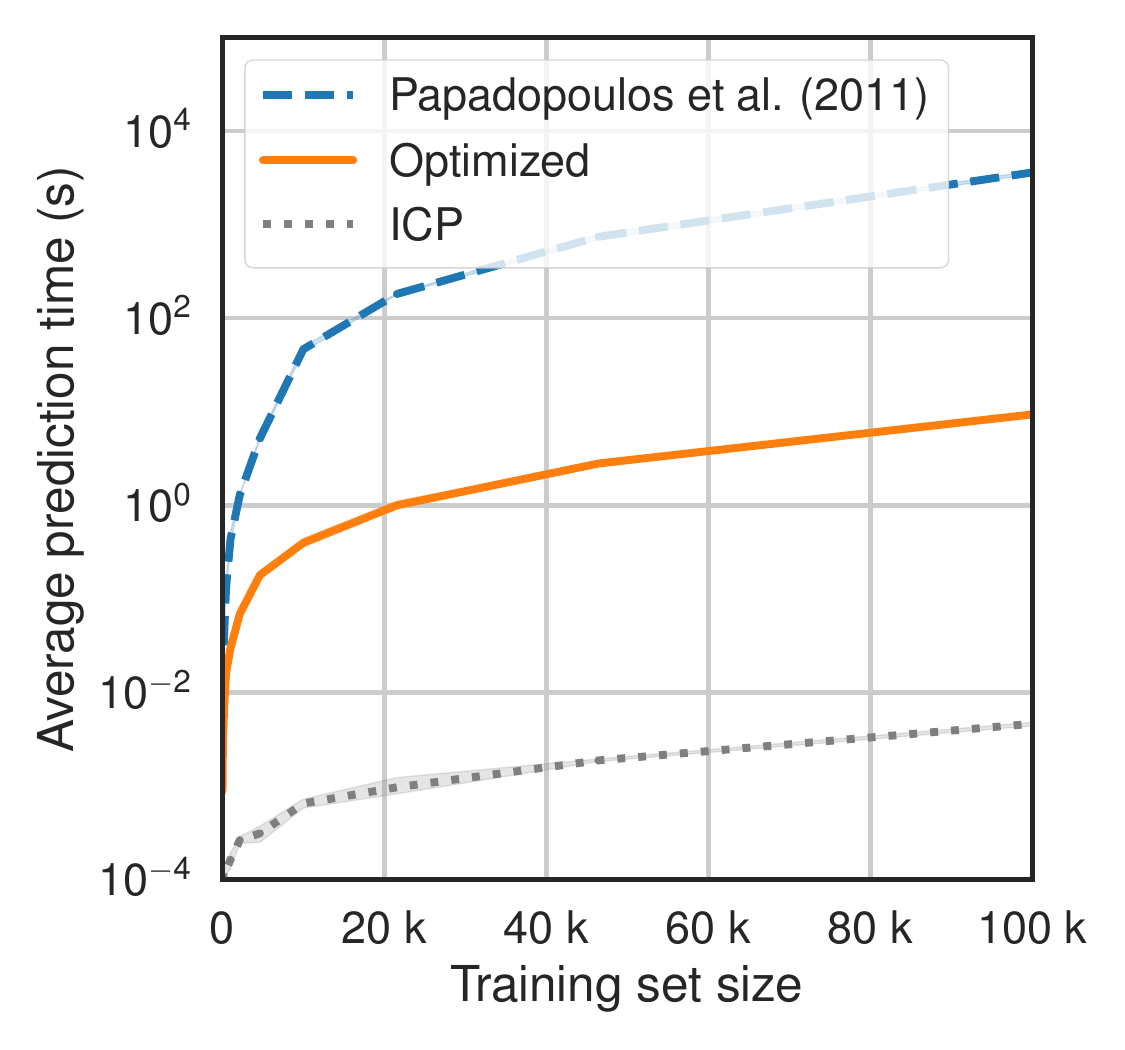}
	\caption{Time comparison of k-NN CP regression: method by \citet{papadopoulos2011regression},
		our optimization, and ICP (baseline).}
	\label{fig:regression}
\end{figure}
\parabf{Empirical evaluation}
We compare full k-NN CP regression \cite{papadopoulos2011regression}
with our optimization
via incremental\&decremental learning.
As a baseline we use ICP k-NN regression, whose
complexity is $\bigoh{tm)}$, where $t \in \{2, ..., n-1\}$ is the
size of the proper training set, and
$m$ the number of test points.
We generate regression examples from $\objspace \times \labelspace = \mathbb{R}^{30} \times \mathbb{R}$
with \texttt{scikit-learn}'s \texttt{make\_regression()} function.
We vary
$n \in [10, 10^5]$, and measure the average prediction
time across $100$ test points.
Each experiment is repeated for $5$ random seeds, and confidence
intervals are plotted.

\autoref{fig:regression} shows that our optimization largely outperforms
the previous version of full k-NN CP regression by \citet{papadopoulos2011regression};
the cost for one prediction with $100$k training points decreases from 1 hour to 9.3 seconds.
ICP outperforms both, taking
roughly 4.6 ms.
We remark, however, that ICP was observed to have a strictly weaker statistical
power in regression \cite{papadopoulos2011regression}.

\parabf{Discussion} We expect that our LS-SVM CP optimization (\autoref{sec:svm})
can be readily applied to speed up the full CP regressor based on ridge regression \cite{nouretdinov2001ridge}.
We leave this, and the optimization of other CP regressors
using incremental\&decremental ideas, to future work.

\section{Discussion and conclusion}
\label{sec:discussion}

Full CP is computationally expensive because of its main routine (\autoref{algo:cp}), which
runs a leave-one-out (LOO) procedure on the ML method
(\textit{nonconformity measure}) that it wraps.
In this paper, we show that if a nonconformity measure can be designed to
learn and unlearn one example efficiently (i.e., it can be trained incrementally\&decrementally),
this can speed up considerably CP classification.
Concretely, we improved k-NN, KDE, and kernel LS-SVM CP classifiers by at least one order magnitude,
and bootstrap CP by a linear factor.
Furthermore, we exploited these ideas to further optimize k-NN CP regression.
Our work makes it feasible to run full CP on large datasets.

We discuss how our optimizations are readily
applicable to other tasks (e.g., clustering, change-point detection), and future directions
for CP optimization.

\parabf{Extensions to more learning tasks}
In addition to classification and regression,
CP is used for tasks such as anomaly detection~\cite{laxhammar2010conformal},
clustering, and sequence prediction \cite{cherubin2016hidden}.
Because all these techniques are based on computing a p-value via
\autoref{algo:cp},
our optimizations are immediately applicable.
For example, conformal clustering \cite{cherubin2015conformal} with k-NN CP
costs $\bigoh{n^2q^p}$, where $q$ is the length of a square grid constructed around $p$-dimensional
training points.
With our optimization, the cost becomes $\bigoh{nq^p}$.
(Usually, $p=2$, by using dimensionality reduction.)

CP has
applications to online learning (e.g., change-point detection \cite{vovk2003testing}).
At step $n+1$, the algorithm trains on examples
$\{(x_i, y_i)\}_{i=1}^n$, makes a prediction for $x_{n+1}$, and learns
the true label $y_{n+1}$.
Adapting our optimizations to this setting is trivial:
it suffice to incrementally learn the new example $(x_{n+1}, y_{n+1})$
after prediction,
which is efficient for
k-NN,
KDE and LS-SVM.
This has a considerable speed-up. For example,
an IID test
by \citet{vovk2003testing},
which has further applications to feature selection~\cite{cherubin2018exchangeability},
requires to incrementally compute a p-value
for the $(n+1)$-th point given $\{(x_i, y_i)\}_{i=1}^n$.
With k-NN CP, this costs $\bigoh{n^3}$; our method
reduces it to $\bigoh{n^2}$ (\autoref{appendix:complexity-derivations}).
Unfortunately, this is not efficient for bootstrap;
we leave its further optimization to future work.

The umbrella of conformal inference also includes
methods such as Venn Predictors (VP),
which give analogous guarantees to CP, but for the calibration of
probabilistic predictions.
Future work may investigate whether VP can be optimized
with similar techniques to the ones we proposed.

\parabf{Boosting and gradient descent}
We hope our work will inspire optimization techniques for more
nonconformity measures.
We foresee as particularly challenging the optimization of methods
such as boosting and gradient descent.
For both techniques, the contribution of a training example depends
on previous examples.
Hence, unlearning an example has a high cost, as it
requires updating the contributions of all the examples that
came after.
We suggest recent work on unlearning methods
may help to achieve this goal.

\parabf{Approximations}
Another natural avenue is to use approximate incremental\&decremental
learning techniques.
For example, by bounding the contribution of each point it
may be possible to achieve very computationally efficient
methods with little cost on statistical efficiency.

\parabf{Exploiting multiple CPUs, GPUs}
A further direction is to study how
to exploit a GPU or multiple CPUs to speed up CP.
Towards this goal, we conducted a preliminary
comparison between parallel and sequential implementations
of CP and optimized CP (\autoref{appendix:multiprocessing});
CP and optimized CP are parallelized in the same way.
Results show that, for a small dataset (5k examples)
standard CP benefits from parallelization, while optimized CP does not
substantially.
Surprisingly, in this case optimized k-NN is even faster without parallelization,
although it does benefit for larger datasets.
More research is needed to determine the
best parallelization strategies for CP, both from an algorithmic
and implementational level.
We leave this, and the study of GPUs for CP, to future work.

\section*{Acknowledgements}
We are grateful to Vladimir Vovk and Adrian Weller for useful discussion.

\bibliography{bibliography}
\bibliographystyle{icml2021}

\clearpage
\newpage
\appendix
\noindent

\onecolumn

\section{Details on CP and CP-inspired methods for classification}
\label{appendix:cp-icp-details}

In the following table, $\nlabels=|Y|$, $\timetrain{n}$ is the training complexity on $n$ examples of the nonconformity measure $\ncm$,
$\timescore{m}$ is its running time for predicting $m$ examples, $n$ and $m$ is respectively the number of
training and test examples.
In ICP and aggregated CP, $t \in \{1, ..., n\}$ is the size of the proper training set.
In cross and aggregated CP, $K \in \{2, ..., n\}$ is a tuning parameter (number of folds).

 \begin{center}
	\begin{tabular}{lcc}
		\toprule
		& Train + Calibrate & Predict \\
		\midrule
		CP   & N/A   & $\bigoh{(\timetrain{n}+\timescore{1})n\nlabels m}$        \\
		ICP & $\bigoh{\timetrain{t}+\timescore{n-t}}$ & $\bigoh{(\timescore{1}+n-t)\nlabels m}$ \\
		Cross CP & $\bigoh{(\timetrain{(\nicefrac{K-1}{K})n} + \timescore{\nicefrac{n}{K}})K}$ &
		$\bigoh{(\timescore{1} + \nicefrac{n}{K})K\nlabels m}$\\
		Aggregated CP & $\bigoh{(\timetrain{t}+\timescore{n-t})K}$ & $\bigoh{(\timescore{1}+n-t)K\nlabels m}$\\
		\bottomrule
	\end{tabular}
\end{center}

\parabf{ICP}
\autoref{algo:icp} describes the algorithms for calibrating
and computing a p-value with ICP.
In the training phase, ICP trains the nonconformity measure $\ncm$ on the
proper training set, $\{(x_1, y_1), ..., (x_{t}, y_{t})\}$,
and uses $\ncm$ to compute the nonconformity scores on
the calibration set $\{(x_{t+1}, y_{t+1}), ..., (x_{n}, y_{n})\}$ (Lines 1-6).
Similarly to CP classification, an ICP classifier computes a
p-value for every $\hat{y} \in \labelspace$ by running
\texttt{COMPUTE\_PVALUE()} (Lines 8-11);
differently from CP, the p-value is only based on the nonconformity
scores computed during calibration and the one for the test object.

\begin{algorithm}[ht!]
	\begin{pseudo}
		\pr{calibrate}(x, y, \{(x_1, y_1), ..., (x_{n}, y_{n})\}, t, \ncm)\\+
		$Z_{train} = \{(x_1, y_1), ..., (x_{t}, y_{t})\}$\\
		$Z_{calib} = \{(x_{t+1}, y_{t+1}), ..., (x_{n}, y_{n})\}$\\
		for $i$ in t+1, ..., n\\+
		$\score_i = \ncm((x_i, y_i); Z_{train})$\\-
		return $\ncm(\cdot; Z_{train}), \{\score_{t+1}, ..., \score_n\}$\\-
		\\
		
		\pr{compute\_pvalue}(x, \hat{y}, \ncm(\cdot; Z_{train}), \{\score_{t+1}, ..., \score_n\})\\+
		
		$\score = \ncm((x, \hat{y}); Z_{train})$\\
		
		$\pval_{(x, \hat{y})} = \frac{\card\{ i=t+1, ..., n \suchthat \score_i \geq \score\} + 1}{n-t+1}$\\
		
		return $\pval_{(x, \hat{y})}$
	\end{pseudo}
	\caption{Inductive Conformal Prediction: computing a p-value for $(x, \hat{y})$}
	\label{algo:icp}
\end{algorithm}

\parabf{CP alternatives}
Algorithms of aggregated CP~\cite{carlsson2014aggregated}, cross CP~\cite{vovk2015cross},
CV+ and jackknife+~\cite{barber2019predictive} can be found in the respective references.
Note that CV+ and jackknife+, albeit originally designed for regression, can be extended to
classification tasks (Appendix D in \cite{gupta2019nested}).

\section{Details on optimized nonconformity measures}
\label{appendix:lssvm-bootstrap-details}

\subsection{LS-SVM}
An LS-SVM model $w$ is learned as a solution of:
$$w = \arg\min_{w \in \mathbb{R}^h} \rho ||w||^2
+ \sum_{i=1}^n (w^\top \phi(x_i)-y_i)^2 \,,$$
for regularization parameter $\rho$.
The closed-form solution to this is:
$$w^* = \Phi[\Phi^\top \Phi + \rho I_n]^{-1}Y \,,$$
where $\Phi = [\phi(x_1), ..., \phi(x_n)]$,
$Y = \{y_1, ..., y_n\}$, and $I_n$ is the identity matrix of size $n \times n$.

The incremental\&decremental learning method by \citet{lee2019exact}
requires storing an auxiliary matrix:
$$C = \Phi[\Phi^\top \Phi + \rho I_n]^{-1}\Phi^\top \,.$$
A description of how to learn incrementally or unlearn an example follows~\cite{lee2019exact}.

\parabf{Incremental learning of one example} To learn $(x_{n+1}, y_{n+1})$, update
the model as follows:
\begin{align*}
w_{new} &= w+\frac{(C-I_q)\phi(x_{n+1})\left(\phi(x_{n+1})^\top w - y_{n+1}\right)}
{\phi(x_{n+1})^\top \phi(x_{n+1}) + \rho - \phi(x_{n+1})^\top C \phi(x_{n+1})}\\
C_{new} &= C + \frac{(C-I_q)\phi(x_{n+1})\phi(x_{n+1})^\top(C-I_q)}
{\phi(x_{n+1})^\top \phi(x_{n+1}) + \rho - \phi(x_{n+1})^\top C \phi(x_{n+1})} \,.
\end{align*}
Where $q$ is the size of the kernel space.

\parabf{Decremental learning of one example} To unlearn $(x_i, y_i)$, update
the model as follows:
\begin{align*}
w_{new} &= w-\frac{(C-I_q)\phi(x_i)\left(\phi(x_i)^\top w - y_i\right)}
				 {-\phi(x_i)^\top \phi(x_i) + \rho + \phi(x_i)^\top C \phi(x_i)}\\
C_{new} &= C - \frac{(C-I_q)\phi(x_i)\phi(x_i)^\top(C-I_q)}
				 {-\phi(x_i)^\top \phi(x_i) + \rho + \phi(x_i)^\top C \phi(x_i)} \,.
\end{align*}

\subsection{Bootstrap algorithm}
\autoref{algo:bootstrap} shows the entire optimized bootstrap CP algorithm.
In the training phase, $B'$ bootstrap samples are generated, and
for some of them (those that do not contain the placeholder ``$\ast$'')
a classifier is trained.
To compute a p-value for example $(x, \hat{y})$, the remaining classifiers
are trained (after replacing ``$\ast$'' with $(x, \hat{y})$), and the
predictions are computed as usual.
A full implementation is provided in the code attached to this submission.

\begin{algorithm}
	\begin{pseudo}
		\pr{train}(\{(x_1, y_1), ..., (x_{n}, y_{n})\}, B)\\+
		$Z^\ast = Z \cup \{\ast\}$ \hspace{3.5cm} // ``$\ast$'' is a placeholder for the test example $(x, \hat{y})$\\
		$E_1, ..., E_n, E \gets \{\}$\\
		\\
		// Associate at least $B$ bootstrap samples to each training example and to ``$\ast$''\\
		for $B'$ in 1, 2, 3, ...\\+
		$Z^\ast_{B'} \gets$ sample $|Z^\ast|$ examples from $Z^\ast$ with replacement\\
		
		for $i=1, ..., n$\\+
			if $(x_i, y_i) \notin Z^\ast_{B'}$\\+
				Insert $Z^\ast_{B'}$ into $E_i$\\--
		if $\ast \notin Z^\ast_{B'}$\\+
			Insert $Z^\ast_{B'}$ into $E$\\-
		
		if $|E| \geq B$ and $|E_i| \geq B$ for all $i=1, ..., n$\\+
		exit for loop\\--
		\\
		// Pretraining for bootstrap samples that do not contain ``$\ast$''\\
		for $i=1, ..., n$\\+
			for $Z_b \in E_i$\\+
				if $\ast \notin Z_b$\\+
					Train classifier $g$ on $Z_b$, and replace element $Z_b$ with
					$g(x)$ in $E_i$\\---
		\\
		// Pretraining for placeholder ``$\ast$''.
		Note: by construction, no element of $E$ contains ``$\ast$''\\
		for $Z_b$ in $E$\\+
			Train classifier $g$ on $Z_b$, and replace element $Z_b$ with $g(x)$ in $E$\\-
		\\
		return $E_1, ..., E_n, E$\\
		\\-
		\pr{compute\_pvalue}((x, \hat{y}), E_1, ..., E_n, E)\\+
		// Compute the nonconformity scores for the training examples\\
		for $i=1, ..., n$\\+
			$\score_i = 0$\\
			for $e \in E_i$\\+
				if $e$ is a pretrained classifier, call it $g(x)$\\+
					$\score_i = \score_i - g^{\hat{y}}(x)$\\-
				else ($e$ is a bootstrap sample $Z_b$ that contains ``$\ast$'')\\+
					Replace ``$\ast$'' with $(x, \hat{y})$ in $Z_b$\\
					Train classifier $g$ on $Z_b$\\
					$\score_i = \score_i - g^{\hat{y}}(x)$\\---
		\\
		// Compute nonconformity score for the test example\\
		$\score = 0$\\
		for $g(x) \in E$\\+
			$\score = \score-g^{\hat{y}}(x)$\\-
		\\
		// Compute p-value\\
		$\pval_{(x, \hat{y})} = \frac{\card \{ i=1, ..., n \suchthat \score_i \geq \score\} + 1}{n+1}$\\
		\\
		return $\pval_{(x, \hat{y})}$
	\end{pseudo}
	\caption{Optimized bootstrap CP algorithm.}
	\label{algo:bootstrap}
\end{algorithm}

\section{Time complexity derivations}
\label{appendix:complexity-derivations}
\subsection{Simplified k-NN and k-NN}

\parabf{Standard}
For simplicity, we only describe the complexity of Simplified k-NN;
the complexity of k-NN is identical to the one derived for
Simplified k-NN up to a linear factor.

Let us define a routine, \texttt{best\_k(A)}, which returns the
$k$ smallest elements of a set $A$ of size $n$.
In our work, we instantiate this to Introselect, which runs
in $\bigoh{n}$ worst-case.\footnote{In our implementation, we base \texttt{best\_k} on \texttt{numpy}'s \texttt{argpartition}.}

The cost for computing the nonconformity measure
$\ncm((x, y); \{(x_1, y_1), ..., (x_{n}, y_{n})\})$
for one example $(x, y)$ requires computing the distances
from $x$ to the training points $\{(x_1, y_1), ..., (x_{n}, y_{n})\}$,
and selecting the $k$ best.
Overall, by using \texttt{best\_k(A)},
this amounts to $\bigoh{n}$.
From the time complexity of CP classification (\autoref{sec:preliminaries}),
we get that running CP with the (Simplified) k-NN nonconformity
measure to predict $m$ test examples takes $\bigoh{n^2\nlabels m}$.

\parabf{Optimized}
In the training phase, we precompute the distances and preliminary scores
for the $n$ training examples ($\bigoh{n^2}$), and store both.
To compute the nonconformity score $\ncm$ for one example $(x_i, y_i)$
in the prediction phase, we only need to compute its distance from
the test example $(x, y)$, and update the provisional score $\alpha_i$
if this distance is smaller than one of the best $k$ distances.
This has cost $\bigoh{1}$.
The cost of CP classification with the optimized measure is
therefore $\bigoh{n\nlabels m}$.

\subsection{KDE}
\parabf{Standard}
Let $\timekernel$ be the time to compute
the kernel on 1 input.
To compute the nonconformity score, we repeat this operation
for all the training points ($\bigoh{\timekernel n}$).
Hence the cost of KDE CP classification is:
$\bigoh{\timekernel n\nlabels m}$.

\parabf{Optimized}
The training costs of this algorithm is $\bigoh{\timekernel n^2}$.
The cost for updating one nonconformity score with our optimization
is $\bigoh{\timekernel}$.
Therefore, the cost of optimized KDE CP for classification
is $\bigoh{\timekernel n\nlabels m}$.

\subsection{LS-SVM}

\parabf{Standard}
The running time of LS-SVM CP is dominated by training LS-SVM.
Let $\bigoh{n^\omega}$, $\omega \in [2, 3]$ be this training cost.
Then, CP classification is $\bigoh{n^{\omega+1}\nlabels m}$.

\parabf{Optimized}
Training the optimized algorithm has the same cost as
training standard LS-SVM, $\bigoh{n^\omega}$.
Let $q$ be the dimensionality of the objects after feature
mapping $\phi(x)$.
To compute the nonconformity score for an example $(x_i, y_i)$,
we: i) update the model with the test example by using the
method by \citet{lee2019exact},
ii) make a prediction for $(x_i, y_i)$.
The first operation has cost $\bigoh{q^3}$), the second one
requires $n$ kernel evaluations and
$\bigoh{q}$ for the dot product.
Therefore, predicting with optimized LS-SVM is $\bigoh{q^3n \nlabels m}$.

\subsection{Bootstrap}

\parabf{Standard}
Let $\timetrain[g]{n}$ be the time needed to train the base
classifier on $n$ training points, and $\timescore[g]{m}$
its running time when computing a prediction for $m$ points.
To compute the bootstrap nonconformity measure once we need
to train $B$ base classifiers and run each one to compute
a prediction. This amounts to $\bigoh{(\timetrain[g]{n} + \timescore[g]{1})B}$.
The overall complexity of bootstrap CP classification is
$\bigoh{(\timetrain[g]{n} + \timescore[g]{1})Bn\nlabels m}$.

\parabf{Optimized}
During the training phase, and for each training point,
we will need to train (and make predictions for), in expectation,
$B\prob{\ast \notin Z_b}$, where
$\prob{\ast \notin Z_b}$ is the probability that an
example (``$\ast$'', in this case) is not contained in
a bootstrap sample of $Z^\ast$ with replacement.
It is easy to see that
$B\prob{\ast \notin Z_b} = B(1-\nicefrac{1}{n+1})^{n+1} \approx Be^{-1}$.
If we repeat the argument for all training examples,
the training phase (Lines 1-26 in \autoref{algo:bootstrap})
takes time $(\timetrain[g]{n} + \timescore[g]{1})Be^{-1}n$.

Computing the p-value for one point
is obtained as the complement of the probability,
$(\timetrain[g]{n} + \timescore[g]{1})B(1-e^{-1})n$.
This is a linear factor $(1-e^{-1}) \approx 0.632$
the speed of the original one.
Overall, CP classification takes 
$(\timetrain[g]{n} + \timescore[g]{1})B(1-e^{-1})n\nlabels m$
for classifying $m$ test examples in $\nlabels$ labels.

\begin{figure}[h]
	\centering
	\includegraphics[width=0.3\textwidth]{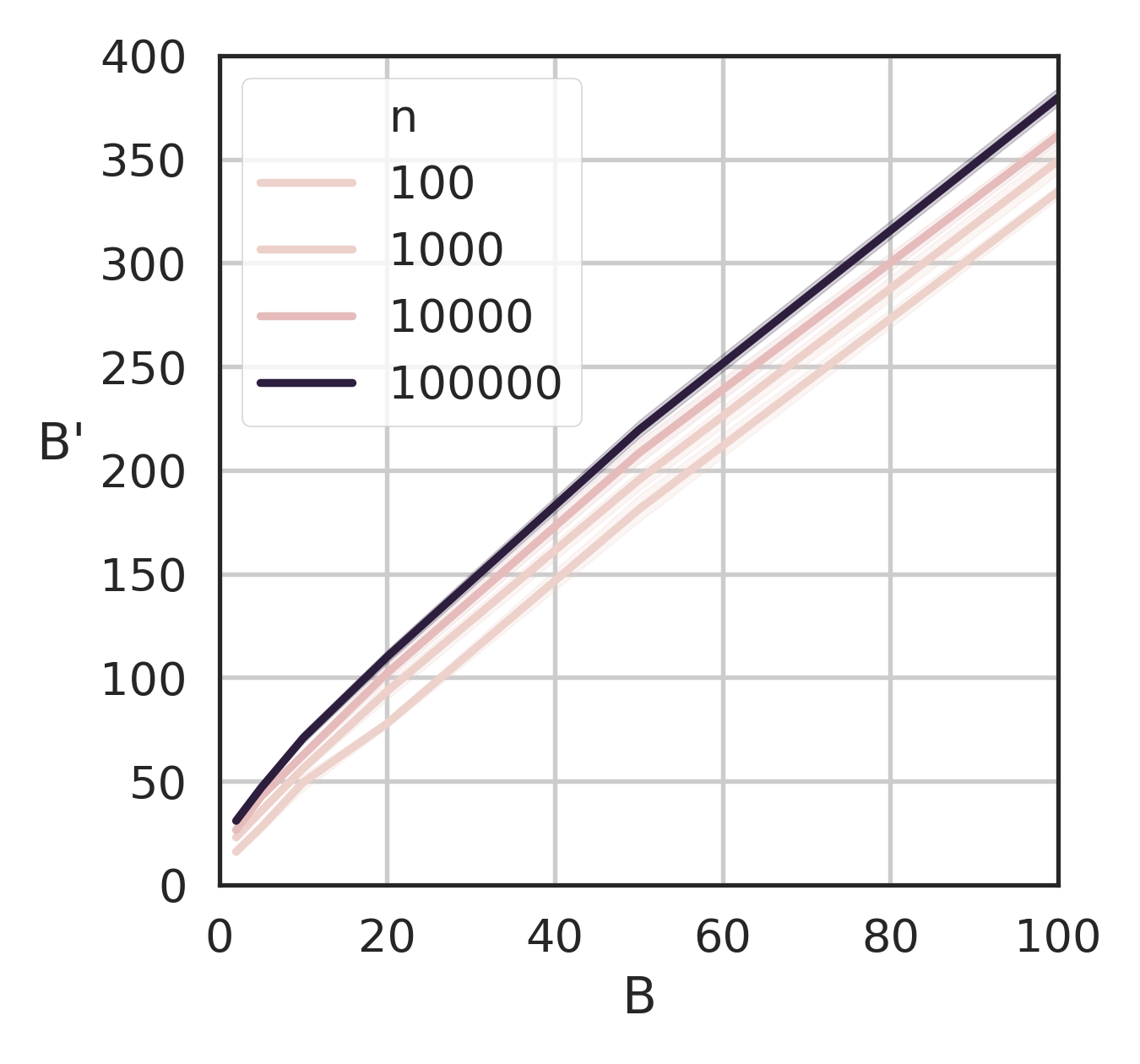}
	\caption{Relation between $B$, $n$, and $B'$ for the optimized bootstrap
		CP algorithm.
	}
	\label{fig:bprime}
\end{figure}

\parabf{Remark}
The actual complexity of optimized bootstrap CP is generally
lower than the one derived above.
Indeed, in the above calculations we assumed that each
bootstrap sample is used for just one point;
however, some bootstrap samples (and respective classifiers)
are in fact shared among several training points.
Therefore, the effective number of classifiers one needs to train
is only $B'$, and not $Bn$.
We show the relation between $B$ and $B'$ in \autoref{fig:bprime},
which indicates that $B' < Bn$.

\subsection{IID test by \citet{vovk2003testing}}

\citet{vovk2003testing} introduced an online algorithm for testing
the exchangeability (or IID-ness) of a sequence of observations.
At step $n$, having observed $\{x_1, ..., x_{n}\}$, it computes a p-value
(\autoref{algo:cp})
for a new observation $x_{n+1}$.
On the basis of the computed p-values, the test derives exchangeability
martingales which can be used as the basis of an hypothesis test.

Suppose we use the k-NN nonconformity measure.
Computing one p-value using standard k-NN CP is $\bigoh{n^2}$.
Since standard CP does not have any way of exploiting previous computations,
the p-values have to be computed independently.
The cost of processing $n$ observations is
$1^2 + 2^2 + ... + n^2 = \sum_{i=1}^n i^2 = \frac{1}{6}n(n+1)(2n+1)$.
Hence, $\bigoh{n^3}$.

When using the optimized k-NN CP, the cost of computing one
p-value for $x_{n+1}$ given $n$ training examples is $\bigoh{n}$;
this includes the cost of training on the new observation.
This means the cost of computing $n$ p-values incrementally, as
required by the IID test, is
$1 + 2 + ... + n = \frac{n(n+1)}{2}$. That is, $\bigoh{n^2}$.

\section{Memory costs}
\label{appendix:memory-costs}
A standard CP implementation requires storing the entire
training data $Z$, which has cost $\bigoh{np}$, where $n$ is the number of examples,
$p$ their dimensionality.
In addition to this cost, the optimized CP versions we introduced
have the following requirements.

\parabf{(Simplified) k-NN} Store, for every training point:
i) the largest among the best $k$ distances,
ii) the provisional score.
This has cost $\bigoh{n}$, negligible w.r.t. the
$\bigoh{np}$ already required by CP.

\parabf{KDE}
Store the $n$ preliminary scores, $\bigoh{n}$,
which is negligible w.r.t. $\bigoh{np}$.

\parabf{LS-SVM} requires storing the
model, $w \in \mathbb{R}^q$, and an auxiliary $q \times q$ matrix
required by the method by \citet{lee2019exact},
where $q$ is the dimensionality of the kernel space.
Including the cost of standard CP, the memory used is
$\bigoh{np + q^2}$.

\parabf{Bootstrap}
In the optimized algorithm, we create $B'$ bootstrap samples.
Of them, on average $B'e^{-1}$ do not contain the placeholder $\ast$;
we can therefore train them and compute a prediction for them
(Lines 16-24).
Storing these predictions has cost $\bigoh{B'e^{-1}}$.
For the remaining ones, we need record the indices pointing
to the augmented dataset $Z^\ast$, so that in the test phase
we can reconstruct the bootstrap samples.
We store $n$ indices for each, totaling a memory
cost of $\bigoh{B'(1-e^{-1})n}$.
The overall cost of bootstrap CP is $\bigoh{np + B'(1-e^{-1})n}$.
The relation between $B'$, $B$, and $n$ is shown in
\autoref{fig:bprime}.

\section{Experiment details}
\label{appendix:experiment-parameters}

\parabf{Hardware and multiprocessing}
We conduct our experiments on a
2x Intel Xeon E5-2680 v3 (48 threads) machine, with
256 GB RAM.
In these experiments,
each time measurement
is performed on a single core.
We limit the number of cores available for the experiment,
so as to prevent time measurements from being affected by other running processes (e.g., kernel tasks).
We prevent \texttt{numpy} from automatically parallelizing
matrix calculations.

\parabf{Hyperparameters}
We use the following hyperparameters for the nonconformity measures.

\begin{table}[h]
	\begin{tabularx}{\linewidth}{lX}
		\toprule
		Method & Hyperparameters\\
		\midrule
		Simplified k-NN \& k-NN & Euclidean distance, $k=15$.\\
		KDE & Gaussian kernel. Bandwidth $h=1$.\\
		LS-SVM & Linear kernel, $\rho = 1$.\\
		Bootstrapping & We instantiate bootstrapping to Random Forest, with
			$B=10$ classifiers. Each classifier, a decision tree, is allowed to
			grow up to depth 10, and to select among $\sqrt{p}$ of features for a split,
			where $p$ is the dimensionality of $\objspace$.\\
		\bottomrule
	\end{tabularx}
\end{table}

When measuring time w.r.t. the training size $n$,
we vary $n$ in the space $[10, 10^5]$,
by evenly separating $13$ values on a log scale.\footnote{Concretely, values for $n$ are obtained
	with \texttt{numpy.logspace(1, 5, 13, dtype='int')}.}

\section{Simplified k-NN results}
\label{appendix:full-results}

\begin{figure}[h]
	\centering
	\includegraphics[width=0.55\textwidth]{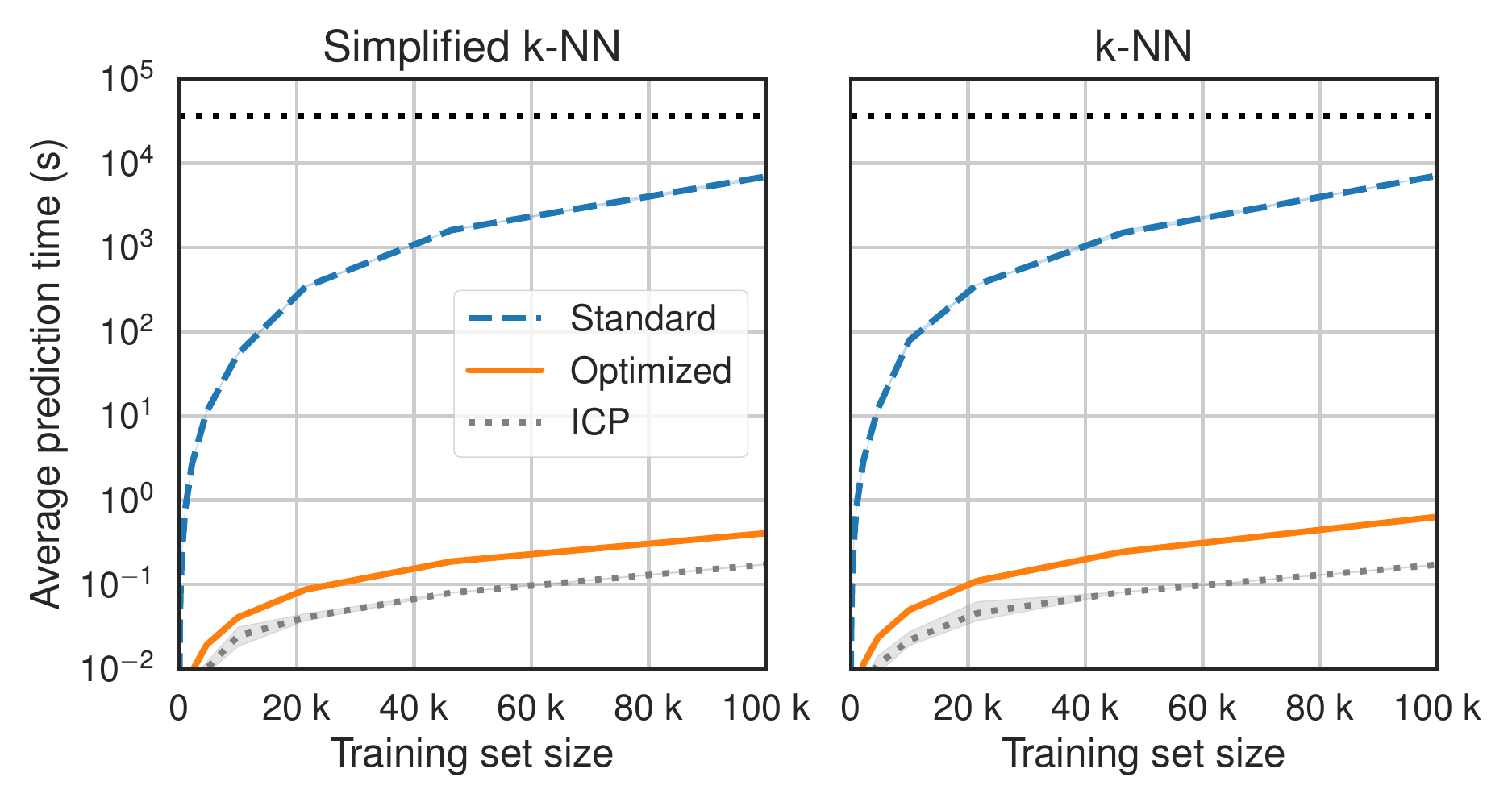}
	\caption{Comparison between standard and optimized k-NN and Simplified k-NN.
		ICP is used as a baseline.
		Prediction time for one test point w.r.t. the size of training data.
		Black dashed line is the  experiment timeout (10 hours).
	}
	\label{fig:full-knn-all}
\end{figure}

Due to lack of space, \autoref{fig:prediction-time-all}
did not include results for Simplified k-NN.
\autoref{fig:full-knn-all} compares k-NN and Simplified k-NN, showing that they
are very similar -- indeed, their asymptotic time complexities are also identical.

\section{Experiments on  \texttt{MNIST}}
\label{sec:mnist}

We conduct experiments on the \texttt{MNIST} dataset~\cite{lecun2010mnist}, which includes 60k training
examples and 10k test examples.
This dataset's records have a much higher dimensionality than
those considered in our previous experiments:
each object is a 28x28 pixel matrix (784 features in total).
Furthermore, this is a 10-label classification setting;
the number of labels strongly penalizes CP (and ICP), although
this is an irreducible cost if we make no assumptions on
$\labelspace$ (\autoref{sec:regression}).
(\citet{fisch2021efficient} recently made some developments w.r.t. this
particular aspect.)

\parabf{Time costs}
We run standard and optimized CP on this dataset, by using
the original training-test split.
We do not include LS-SVM in this set of experiments,
as it is specific to binary classification (although
it could be extended by using e.g. a one-vs-all approach).

\begin{table}
	\centering
\begin{tabular}{llll}
	\toprule
	{}            & CP       & Optimized CP     & ICP              \\ \midrule
	NN            & 0s / T(1) & 34m 5s / 7h 9m   & 22m 58s / 2h 38m \\
	SimplifiedKNN & 0s / T(1) & 29s / 4h 36m     & 3m 47s / 1h 38m  \\
	KNN           & 0s / T(0) & 34m 17s / 7h 21m & 20m 25s / 2h 29m \\
	KDE           & 0s / T(1) & 1h 17m / 29h 13m & 1h 30m / 6h 13m  \\
	RandomForest  & 0s / T(0) & 48s / T(0)        & 21s / 1h 25m     \\ \bottomrule
\end{tabular}
\caption{\texttt{MNIST} evaluation. Each entry reports \textit{training/prediction} time for 60k training and
	10k test points.
	$T (p)$ indicates that the timeout (48h) was reached before completing all the predictions,
	and that $p$ predictions were made by then.}
\label{tab:mnist}
\end{table}

\autoref{tab:mnist} indicates that the advantage of using
optimized CP w.r.t. standard CP is substantial; in particular,
for the fastest nonconformity measures, standard CP could make just
1 prediction (out of 10k test points) within the 48h timeout limit.
Results also suggest that exact optimized CP is a
practical alternative to ICP; for example, optimized Simplified k-NN
CP run in 4.3 hours, while ICP with the same nonconformity measure
run in 1.6 hours.
Unfortunately, optimized Random Forest was unable to make any
predictions within the 48h timeout; we hope this method
can be further optimized in the future.
Finally, we observe that optimized CP KDE was substantially worse
than ICP; the reason is that, for the experiments on \texttt{MNIST},
we used arbitrary precision math
to make KDE numerically stable -- something that can be
improved upon.

\parabf{Statistical Efficiency of CP and ICP}
As a byproduct of the above experiment, we are able to
compare CP and ICP on the basis of their statistical power
(efficiency).
Note that an analysis on such a large dataset
would not have been possible without using our CP optimizations.

We compare CP and ICP in terms of their \textit{fuzziness}
on the \texttt{MNIST} test set.
The fuzziness of a set of p-values $\{\pval_{(x, y)}\}_{y\in Y}$,
returned by CP or ICP as the prediction for test object $x$, is the average
of the p-values excluding the largest one:

$$\sum_{y\in Y} \pval_{(x, y)} - \max_y \pval_{(x, y)} \,.$$

A smaller fuzziness indicates better performance~\cite{vovk2016criteria}.
We use a Welch one-sided statistical test for the
null hypothesis $H_0$: ``ICP has a smaller fuzziness (i.e., it is better) than CP''.
We reject the null hypothesis for a p-value $< 0.01$.

The table below indicates the fuzziness of the evaluated techniques
for the \texttt{MNIST} dataset. An asterisk \textbf{*} indicates statistical significance.
Random Forest CP was excluded, as it did not return predictions
within the timeout (\autoref{tab:mnist}).

Results demonstrate CP is consistently better w.r.t. fuzziness
than ICP.
However, we observe that future work is needed to compare
CP and ICP under various conditions (e.g., umbalanced data,
distribution shift, ...). Our optimizations make this
analysis feasible.

\begin{center}
	\begin{tabular}{lcc}
		\toprule
		{} & CP & ICP\\
		\midrule
		NN            &  \textbf{0.00047 $\pm$ 0.00105*} &  0.00065 $\pm$ 0.00143 \\
		Simplified k-NN &      \textbf{0.04998 $\pm$ 0.07151*} &     0.05684 $\pm$ 0.07744\\
		k-NN & \textbf{0.00066 $\pm$ 0.00125*} & 0.00098 $\pm$ 0.00174\\
		KDE           &      \textbf{0.04494 $\pm$ 0.07005*} &      0.16791 $\pm$ 0.11729\\
		\bottomrule
	\end{tabular}
\end{center}

\section{Does multiprocessing help?}
\label{appendix:multiprocessing}

We consider a multiprocessing implementation for CP
classification.
The CP implementation parallelizes \autoref{algo:cp},
which is run for every label and test point.
Standard and optimized CP are parallelized in the same way.
For the parallel versions, we employ a Python \texttt{Pool}
of processes, as shown in the code included in the supplementary
material.

We used the 48 threads machine described
in \autoref{sec:experiments}, and made all the cores available
for multiprocessing.
We generated a dataset of size 1000, split it into
training (\%70) and test sets, and timed
the sequential and parallel versions.
Due to the high complexity of standard CP,
we could not evaluate this for larger datasets.
We report the measurements collected over 5 runs.

Results (\autoref{tab:parallel-1000})
indicate that CP with standard nonconformity measures
always benefits from parallelization, bringing
at least one order of magnitude speed up.
Conversely, optimized nonconformity measures give
a mixed picture:
except for LS-SVM and Random Forest, parallelization
only brings mild improvements.
Surprisingly, optimized k-NN CP is faster than the respective
parallel version.

After this observation,
we repeated the experiment just for optimized k-NN CP,
for a dataset of 100k examples.
In this case, parallelization indeed helps:
prediction takes 1 hour for the parallel version,
3.5 hours for the sequential one.
We conclude that the benefit of multiprocessing
exists, but only for large datasets.
We suspect this is due to the overhead of
creating new processes, and that it can be further
optimized in the future by working on the implementational details.

\begin{table}[ht!]
	\caption{Time comparison of sequential and parallel implementations, based on a dataset of
		1000 examples with $30$ features each. Time is measured in seconds.}
	\label{tab:parallel-1000}
	\centering
	\begin{tabular}{clcc}
		\toprule
		{} & {} &     CP &  CP Parallel  \\
		\midrule
		\parbox[t]{2mm}{\multirow{6}{*}{\rotatebox[origin=c]{90}{Standard}}}
		&Simplified k-NN   &   74.60 $\pm$ 1.46  & \textbf{  3.95 $\pm$ 0.18 } \\
		&k-NN     &   82.14 $\pm$ 1.29  & \textbf{  4.11 $\pm$ 0.17 } \\
		&KDE     &   138.48 $\pm$ 1.49  & \textbf{  6.66 $\pm$ 0.28 } \\
		&LS-SVM     &  8852.18 $\pm$ 27.46  & \textbf{ 624.74 $\pm$ 2.58 } \\
		&Random Forest   &  5061.67 $\pm$ 310.75  & \textbf{ 225.75 $\pm$ 4.41 } \\
		\midrule
		\parbox[t]{2mm}{\multirow{6}{*}{\rotatebox[origin=c]{90}{Optimized}}}
		&Simplified k-NN  &   1.29 $\pm$ 0.01  & \textbf{  0.43 $\pm$ 0.05 } \\
		&k-NN    & \textbf{  1.60 $\pm$ 0.27 } &   9.04 $\pm$ 0.33  \\
		&KDE    &   1.35 $\pm$ 0.14  & \textbf{  0.41 $\pm$ 0.00 } \\
		&LS-SVM    &   8.47 $\pm$ 0.58  & \textbf{  0.76 $\pm$ 0.01 } \\
		&Random Forest  &  2409.59 $\pm$ 546.21  & \textbf{ 110.32 $\pm$ 32.50 } \\
		\bottomrule
	\end{tabular}
\end{table}

\end{document}